\documentclass[10pt,twocolumn,a4paper]{esaAI}

\usepackage{draftwatermark}
\SetWatermarkText{Accepted Pre-print}
\SetWatermarkScale{0.5}

\usepackage{algorithm}
\usepackage{algpseudocode}
\usepackage{xcolor}

\begin{document}

\title{AI-Driven Risk-Aware Scheduling for Active Debris Removal Missions}

\def\authorEmail{adam.abdin@centralesupelec.fr}

\author[1]{Antoine Poupon\footnote{Equal contributions.} }
\author[1]{Hugo de Rohan Willner$^*$}
\author[1]{Pierre Nikitits$^*$}
\author[2]{Adam Abdin \thanks{Corresponding author. E-Mail: \authorEmail}}

\affil[1]{Université Paris-Saclay, CentraleSupélec, Gif-sur-Yvettes, France}
\affil[2]{Université Paris-Saclay, CentraleSupélec, Laboratory of Industrial Engineering, Gif-sur-Yvette, France}

\makeCustomtitle

\begin{abstract}
The proliferation of debris in Low Earth Orbit (LEO) represents a significant threat to space sustainability and spacecraft safety. Active Debris Removal (ADR) has emerged as a promising approach to address this issue, utilising Orbital Transfer Vehicles (OTVs) to facilitate debris deorbiting, thereby reducing future collision risks. However, ADR missions are substantially complex, necessitating accurate planning to make the missions economically viable and technically effective. Moreover, these servicing missions require a high level of autonomous capability to plan under evolving orbital conditions and changing mission requirements. In this paper, an autonomous decision-planning model based on Deep Reinforcement Learning (DRL) is developed to train an OTV to plan optimal debris removal sequencing. It is shown that using the proposed framework, the agent can find optimal mission plans and learn to update the planning autonomously to include risk handling of debris with high collision risk. 

\end{abstract}

\section{Introduction}

In recent years, Low Earth Orbit (LEO) has become increasingly crowded with debris originating from spacecraft fragmentation at the end of their operational life cycle. Collisions of such debris with space assets can often lead to catastrophic failure.  One example is the Cosmos 2251-Iridium 33, which led to the destruction of the Iridium 33 satellite and added more than 2000 pieces of debris to LEO \cite{nicholas2009collision}. In addition, collisions between objects generate new space debris, increasing the likelihood of further collisions, in what is known as the Kessler effect \cite{kessler2010kessler_syndrome}.  Although international post-mission disposal policies are being developed, to this date, they are not sufficient to handle this growing problem \cite{liou2013update}. 

To mitigate debris growth, studies recommend a removal rate of five heavy debris per year \cite{flury2001updated}. Many approaches have been proposed to remove or ``deorbit'' these debris. A promising method is Active Debris Removal (ADR). ADR missions are performed by an Orbital Transfer Vehicle (OTV), which can visit and de-orbit multiple debris during a single mission. 
Due to high costs, ADR missions should be able to maximise debris removal per mission at minimal cost. However, finding the optimal planning for de-orbiting several debris is non-trivial and requires adapted tools to find the most adequate plans. In addition, these missions require a high level of autonomous planning capabilities for the robotic spacecraft to be able to adapt to changing orbital conditions and mission requirements rapidly and effectively.

The ADR mission planning problem can be formulated as a Cost-Constrained Traveling Salesman Problem (CCTSP) \cite{sokkappa1991cost}, a variant of the Traveling Salesman Problem (TSP) \cite{gavish1978travelling}. Key constraints are the mission duration and the propellant available to the OTV.  As the OTV transfers from one debris to the next, it consumes both time and propellant. Each debris is given a value, allowing the mission designer to prioritise for a certain goal, such as debris size \cite{yang2019rl_for_adr} or collision risk. The objective is, therefore, to optimise the debris sequence for maximum mission value under these constraints.

Traditional optimisation models of ADR missions focus on minimising the mission cost using static methods. However, recent AI advancements, particularly in reinforcement learning (RL), allow the consideration of a dynamic approach. RL allows sequential decision-making and adaptability to new information \cite{ai2022deep}. Applying RL to ADR mission planning is relevant since OTVs receive ongoing monitoring data and may need to update plans accordingly, ensuring their capacity to operate autonomously. 

Research in ADR mission planning varies mainly in transfer strategy and optimisation approach. Ion thrusters for Low-Thrust transfers and chemical propulsion for High-Thrust transfers have been studied using various static optimisation methods \cite{zuiani2012preliminary}\cite{medioni2023trajectory}. RL frameworks, such as Deep Q-Network (DQN) \cite{mnih2013playing}, have shown promise in optimising debris selection dynamically, but their adaptive capabilities need further exploration \cite{yang2019rl_for_adr}\cite{yang2020ucbts}.

This paper aims to explore the adaptive capabilities of an RL-controlled ADR mission. The initial objective is to create an RL environment for High-Thrust ADR missions. Subsequently, the study shows that an agent can be trained to autonomously respond according to new information obtained during the mission.

\section{Methods}

\subsection{General framework for ADR missions}

In an ADR mission, a robotic OTV transfers from one debris to another, captures it, and deorbits each debris until it exceeds its mission time or allotted propellant. Given that the OTV's resources only allow it to visit a subset of debris within a larger debris set, the goal is to find the optimal sequence of debris to visit to maximise mission value under technical and economic constraints. In this work, the mission value is associated with the debris collision risk.

Using the Keplerian reference frame, and assuming circular orbits, a transfer from one debris to the next is represented as: ($a_1$, $i_1$, $\omega_1$, $\nu_1$) $\rightarrow$ ($a_2$, $i_2$, $\omega_2$, $\nu_2$).

\noindent where ($a$), ($i$), ($\omega$) and ($\nu$) represent the semi-major axis, inclination, argument of periapsis and true anomaly, respectively.
Transfer strategies can be a sequence of individual manoeuvres (e.g., ($a_1$) $\rightarrow$ ($a_2$), ($i_1$) $\rightarrow$ ($i_2$)), or combined manoeuvres (e.g., ($a_1$, $i_1$) $\rightarrow$ ($a_2$, $i_2$)).

\subsection{Reinforcement Learning formulation}

The ADR Mission planning problem can be formulated as a Markov Decision Process (MDP) to be solved using RL. The main components of the proposed MDP framework are shown in this section.
\subsubsection{State} \label{sec:State}
Inspired by \cite{yang2019rl_for_adr}, in this formulation, the state is represented as a vector shown in Table (\ref{table:state_vector}). Element 1 is the number of debris left to remove. Element 2 is the total fuel left for the OTV, calculated as the remaining possible ($\Delta V$). Element 3 is the total mission time left ($\Delta T_{\text{left}}$). Element 4 is the current debris to be removed. Element 5 is a binary flag that describes the ``removed or not'' status of each debris. Element 6 is the collision risk associated with each debris. Each piece of debris is assigned a collision risk level ranging from 1 to 10, indicating its deorbiting priority, which can change during the mission. All possible states ($s$) construct the state space ($S$). Both the length of the Removal Flags and collision risk lists are equal to the amount of debris considered for deorbiting ($N$).

\begin{table*}[h]\renewcommand{\arraystretch}{1.2}
\small
\begin{center}
\begin{tabular}{c || c | c | c | c | c | c | c} 
\hline\hline
Element &  $N_{\text{debris left}}$&  $\Delta V_{\text{left}}$&  $\Delta T_{\text{left}}$&  Current Location&  Removal Flags& collision risk\\
\hline\hline
Example &  4&  1200 m/s&  2 days&  Debris 2&  \{0,1,0,1\}& \{1,10,1,1\}\\\hline\hline
\end{tabular}
\caption{Example of state vector.}
\label{table:state_vector}
\end{center}
\end{table*}

\subsubsection{Action}

At each step, the OTV captures debris. The agent's action is to select the next debris to deorbit. The action space ($A$) is therefore constructed of all actions $a = d$, where $\left\{d \in \mathbb{N}^{+} \mid 1 \leq d \leq N\right\}$ is the order of debris in the list ($N$).
It is assumed that the time of arriving at the target debris is also the time of leaving for the next one (ignoring the time of the de-orbiting maneuver) as it is negligible compared to orbital transfer times. Therefore,this action formation without a waiting procedure is applicable.

\subsubsection{State Transition}\label{sec:State_Trans}

Unlike traditional RL methods in a control context where the state transition is associated with a time step, the formulation proposed in this work associates the state transition with a \emph{removal step}. Hence, given a state and action, the next state is computed using the state transition function described in Eq.({\ref{transition_function}})
\begin{align}
\label{transition_function}
    \left\{\begin{array}{l}
    s^{\prime}[N_{\text{debris left}}]=s[N_{\text{debris left}}]-1 \\
    s^{\prime}[\Delta V_{\text{left}}]=s[\Delta V_{\text{left}}]-\Delta V_i \\
    s^{\prime}[\Delta T_{\text{left}}]=s[\Delta T_{\text{left}}]-\Delta T_i \\
    s^{\prime}[Current Location] = a \\
    s^{\prime}[Removal Flags][a] = 1 \\
    s^{\prime}[Collision Risk] = RandRisk(s^{\prime}[Removal Flags])
    \end{array}\right.
\end{align}
After taking an action, the state transition updates the number of debris left, the fuel and time left for the rest of the mission, the current location of the OTV, the binary flag of the last deorbited debris, and the collision risk of all debris.

To validate the proposed framework on a case study, a simple risk setting function ``$RandRisk$'' is used. It randomly assigns a high collision risk to available debris according to the logic shown in Algorithm (\ref{alg:randRisk}).

\begin{figure}[H]
  \centering
  \begin{minipage}{.9\linewidth}
    \begin{algorithm}[H]
      \caption{RandRisk}
      \label{alg:randRisk}
        \begin{algorithmic}[1]
            \fontsize{7}{9.6}\selectfont
            \State $Risk\_list \gets \text{Reset all to 1}$
            \State $available\_debris \gets \text{list of indices i where binary\_flags[i] is 0}$
            \If{$random() < Threshold$}
                \State $Risk\_debris \gets \text{random choice from } available\_debris$
            \EndIf
            \State $Risk\_list[Risk\_debris] \gets R_\text{prio}$
            \State \Return $Risk\_debris$
        \end{algorithmic}
    \end{algorithm}
  \end{minipage}
\end{figure}

Where ($R_\text{prio}$) is the reward the agent will receive for deorbiting a high-risk debris. 

It is important to note that, in this algorithm, the ``$RiskLevel$'' of all debris is reset to 1 after every removal step. This ensures that $R_\text{prio}$ is only available for a single step, as further explained in Sec. (\ref{sec:reward}). 

\subsubsection{Reward} \label{sec:reward}

The reward function is designed to ensure that the OTV minimises the overall risk in the system by deorbiting debris according to their risk level. To do so, a \emph{deterministic} reward function is used, as described in Eq. \ref{eq:reward_function}.

\begin{equation}
    r(s,a) = 
    \begin{cases}
    s[Collision Risk][a] & \text{if } \text{($s$) is a non-terminal} \\
    & \text{state} \\
    0 & \text{if } \text{($s$) is a terminal state}
    \end{cases}
\label{eq:reward_function}
\end{equation}

The agent receives a positive reward proportional to the collision risk of the debris removed, and no reward if no debris is removed. Moreover, since $RiskLevel$ is reset at every removal step ($i$), the agent only receives a high reward if it takes the action ($a$) that deorbits the high-risk debris immediately after receiving the risk information in state ($s$).

\subsubsection{Terminal States}

The terminal states, which represent the end of a training episode, are used in the algorithm in two different scenarios. First, if the agent takes an action that results in a state where it has exceeded its allocated resources ($\Delta V_{\text{max}}$) or ($\Delta T_{\text{max}}$), and second, if the agent takes an action that is impossible (e.g., revisiting debris already deorbited). These terminal states are coupled with a reward of 0.

\subsubsection{Orbital Transfer Simulation}

A high-thrust transfer simulator is developed to calculate the orbital transfer times and fuel consumption of the OTV during the learning process. It takes the action selected by the agent (i.e., the next debris to visit) and outputs the time of the maneuver and the fuel cost. The simulator is based on the use of 3 sequential maneuvers to transfer from the current debris to the next.

\subsection{DQN Agent}

A Deep Q-Network architecture is used to train the agent, given its ability to handle the continuous state-space of the MDP model described in Sec. (\ref{sec:State}). 

The network takes as input the state vector described in (\ref{sec:State}) and outputs Q-values, which represent the estimated value of deorbiting each debris given the current state. These values guide the agent's decision-making process during the mission. The function approximation network learns the optimal Q-values based on the reward sampled from the environment at each step, by minimising the difference between the Q-value prediction and the Q-target described in Eqs. (\ref{eq:q_value_prediction}) and (\ref{eq:q_value_loss}).

\begin{align} \text{Q-value prediction: } &
    y_i = \begin{cases}
    R_i \quad \text{if episode} \\
    \quad \quad \text{terminates at step i+1} \\
    R_i + \gamma \max_{a'} Q(s_{i+1}, a'; \theta_{target}) \\
    \quad \quad \text{otherwise}
    \end{cases} \label{eq:q_value_prediction}\\
    \text{Q-value loss: } \quad & (y_i - Q(s_i, a; \theta))^2 \label{eq:q_value_loss}
\end{align}
where \textit{(i)} is the removal step and ($s_i$), ($a_i$) and ($R_i$) are respectively the state, the action and the reward at removal step ($i$).

The neural network used in this paper has a straightforward  architecture consisting of two hidden layers with ReLU activation functions. As discussed in \cite{mnih2013playing}, a Fixed Q-Target approach was used to stabilise the learning. This is done by implementing separate Value and Target networks, parameterised by ($\theta_{value}$) and ($\theta_{target}$), respectively. 

Finally, since sampling from the environment is computationally expensive, a \emph{Replay\_Buffer} is used to make a more efficient use of experiences, and reduce the correlation between experiences.

\section{Results}

To benchmark the agent on realistic data, the Iridium-33 dataset was used, containing the coordinates of 320 debris resulting from collisions between the satellites Cosmos 2251 and Iridium 33. Hyperparameter tuning of the agent was undertaken on this dataset using Bayesian Optimisation \cite{snoek2012practical}.

\subsection{Validation of the DRL results}

To validate the RL algorithm's effectiveness, an exhaustive search is performed on a small subset of debris, and used as a baseline for comparison with the RL agent's mission. The full depth search results are known to be optimal, hence this allows the comparison of the \emph{theoretical} maximum reward and the reward received by the RL agent.

\begin{figure}[htbp] 
    \centering 
    \includegraphics[width=.95\columnwidth]{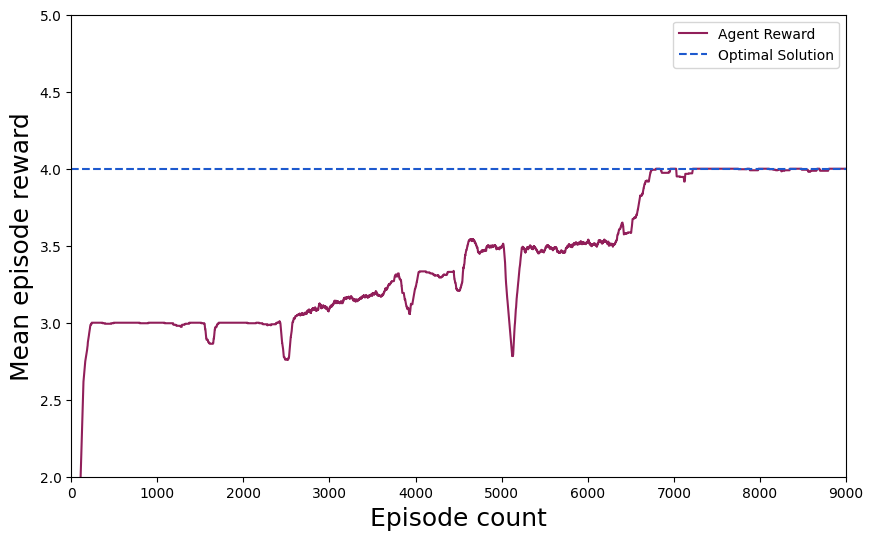}
    \caption{RL Agent Validation - the agent converges to the theoretical optimal solution.}
    \label{fig:RL Agent Validation}
\end{figure}

Given the available resources ($\Delta V_{\text{max}}$) or ($\Delta T_{\text{max}}$), and a debris dataset, a full-depth search algorithm is designed to determine the theoretical maximum sequence length. Since the computational cost of the exhaustive search increases exponentially, this full-depth algorithm can only be used on small datasets for validation. To further scrutinise the RL agent, ($\Delta V_{\text{max}}$) is chosen so that the maximal solution subset is of size one, i.e., only a single debris sequence can achieve the maximum reward. Finally, since the aim is to validate the maximal sequence length, the environment is set without considering collision risks.
\\
The validation methodology is the following:\\
\indent (1) Using a debris list length of $N=10$. Compute all of the possible subsets of size $N_\text{subset}=5$, i.e., all possible sequences of debris of length 5.\\
\indent (2) Using the orbital transfer simulator, compute the ($\Delta V_{\text{total}}$) for each subset of debris visited.\\
\indent (3) Find the subset that used the minimum ($\Delta V_{\text{total}}$),  such that   $min(\Delta V_{\text{total}}) = \Delta V_{\text{optimal}}$.  \\
\indent (4) Run the RL algorithm on the same $N=10$ debris, with a constraint of  $\Delta V_{\text{max}}= \Delta V_{\text{optimal}}$.\\
\indent (5) Verify that the agent converges to the optimal sequence. \\
Following this methodology, agents are trained over different initiating seeds and the results of the DRL obtained solutions are compared to the guaranteed optimal solutions found via exhaustive search. The agent learning progress over training episodes is shown in Fig. (\ref{fig:RL Agent Validation}).

Fig.~\ref{fig:RL Agent Validation} shows the average episode reward as a function of the training episodes. As seen in the figure, the results confirm the agent's ability to find the theoretically optimal solution.

\subsection{Case Study Results}

To evaluate the agent's performance, two scenarios are tested. In both, the agent receives a high reward for deorbiting high-risk debris. In the first scenario, the agent does not have access to collision risk information and thus receives the high reward only if it happens to select and deorbit high-risk debris by chance. This serves as a baseline for comparison with the second scenario, where the agent has direct access to collision risk levels.

\begin{figure}[htbp] 
    \centering 
    \includegraphics[width=.95\columnwidth]{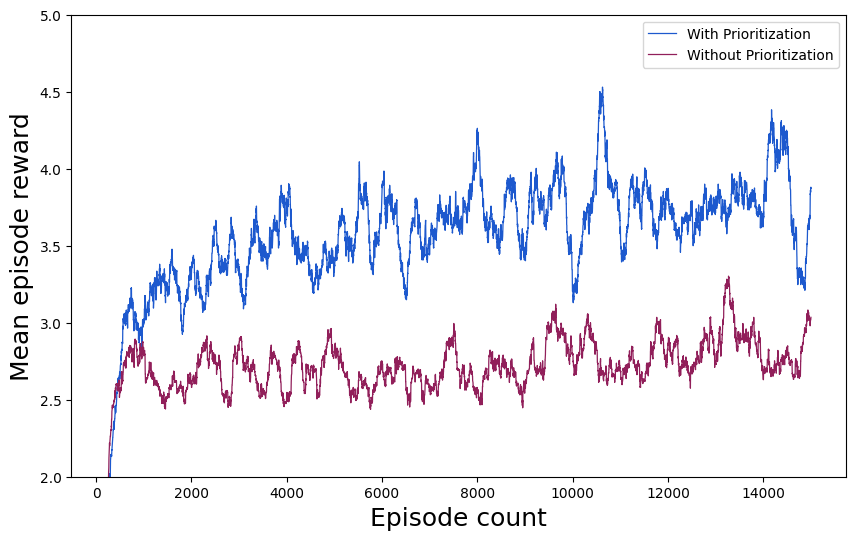}
    \caption{RL Agent performance - comparison between having access vs no-access to collision risk estimations.}
    \label{fig:prio_access_comp}
\end{figure}

Fig.~ (\ref{fig:prio_access_comp}) compare the two scenarios over multiple initialisation seeds. It can be seen that the agent with access to risk level information successfully learns to exploit them, resulting in a higher average reward over mission time. These results confirm the agent's ability to react to collision risk information given mid-mission by adjusting its planning autonomously for which debris to visit.

\section{Discussion}

In this work, an RL-based autonomous planning algorithm for OTV is developed for deorbiting the maximum amount of debris in a single mission, considering a limited amount of fuel and mission time. The results show that this algorithm is capable of finding optimal sequence of deorbiting debris. Moreover, it is shown that using the proposed framework, the agent can learn to update the planning mid-mission based on collision risk estimations, with the aim of reducing the overall risk in the system.

\printbibliography
\addcontentsline{toc}{section}{References}

\end{document}